\documentclass{article}

\PassOptionsToPackage{numbers, square}{natbib}
\PassOptionsToPackage{table,xcdraw,dvipsnames}{xcolor}
\usepackage{floatrow}%

\usepackage[accepted]{icml2024}
\usepackage{amsmath}
\usepackage{stackengine} 
\stackMath
\usepackage{graphicx}

\usepackage{caption}
\usepackage{multirow}

\usepackage[utf8]{inputenc} 
\usepackage[T1]{fontenc}    
\usepackage{hyperref}       
\usepackage{url}            
\usepackage{booktabs}       
\usepackage{amsfonts}       
\usepackage{nicefrac}       
\usepackage{microtype}      
\usepackage{adjustbox}
\usepackage{subcaption}
\icmltitlerunning{Revealing the Utilized Rank of Subspaces of Learning in Neural Networks}
\begin{document}

\twocolumn[
\icmltitle{Revealing the Utilized Rank of Subspaces of Learning in Neural Networks}



\icmlsetsymbol{equal}{*}

\begin{icmlauthorlist}
\icmlauthor{Isha Garg}{yyy}
\icmlauthor{Christian Koguchi}{yyy}
\icmlauthor{Eshan Verma}{yyy}
\icmlauthor{Daniel Ulbricht}{yyy}
\end{icmlauthorlist}

\icmlaffiliation{yyy}{Apple}

\icmlcorrespondingauthor{Isha Garg}{i\_garg@apple.com}

\icmlkeywords{Machine Learning, ICML}

\vskip 0.3in
]



\printAffiliationsAndNotice{}  




\begin{abstract}
In this work, we study how well the learned weights of a neural network utilize the space available to them. This notion is related to capacity, but additionally incorporates the interaction of the network architecture with the dataset. Most learned weights appear to be full rank, and are therefore not amenable to low rank decomposition. This deceptively implies that the weights are utilizing the entire space available to them. We propose a simple data-driven transformation that projects the weights onto the subspace where the data and the weight interact. This preserves the functional mapping of the layer and reveals its low rank structure. In our findings, we conclude that most models utilize a fraction of the available space. For instance, for ViTB-16 and ViTL-16 trained on ImageNet, the mean layer utilization is 35\% and 20\% respectively. Our transformation results in reducing the  parameters to 50\% and 25\% respectively, while resulting in less than 0.2\% accuracy drop after fine-tuning. We also show that self-supervised pre-training drives this utilization up to 70\%, justifying its suitability for downstream tasks.
\vspace{-2mm}
\end{abstract}

\section{Introduction}
\label{sec:intro}

The notion of `capacity' of a network becomes less clear as we scale to large, deep neural networks. In practice, it is often thought of as a function of the number of parameters in the network. 
In this work, we shift out attention to the concept of \textit{utilization}, which we define distinctly from model capacity in that it captures the interaction between both the complexity of a trained network and the dataset its trained on. 
We address utilization from a subspace perspective. Most learned weights appear to be full rank, suggesting we cannot trivially perform a low rank decomposition. In this work, we show that only a fraction of these dimensions interact with the data the weight operates on.
We study the low rank decomposition of the input and output to the layers rather than the weights directly and find a simple modification that preserves the layer mapping by projecting the weight onto the subspaces of interaction. We refer to this as the \textit{effective subspace} where learning occurred, and the dimension of this subspace as the \textit{utilized rank} for that layer. This lower dimensional subspace allows for easy decomposition and efficiency by reducing the number of parameters and FLOPs. It also allows us to compare different networks in terms of their \textit{Mean Layer Utilization} (MLU), a statistic that is informational for studying the structure of networks.


 Suppose the input and output for a given layer live on subspaces $S$ and $T$ respectively. Then, projecting the input onto $S$ and the output onto $T$ is invariant in the forward pass up to some allowable $L_2$ error. We show that performing these two projections is similar to performing the forward pass with a transformed weight matrix $W$, with its row space projected onto $S$ and its column space projected onto $T$. We upper-bound the error resulting from this transformation, and show that it can be driven down by controlling the spectral energies of the input and output subspaces. This transformation reveals the \textit{utilized rank} of $W$, which we find to be far lower than the intrinsic rank of the original $W$. We determine the rank for a single layer by performing a binary search over the singular values of $S$ and $T$ to limit the resulting error from this transformation on the validation set.  This allows us to find the utilized ranks of all layers without retraining, with a predictable and bounded accuracy drop that can easily be recovered via finetuning. 

Studying the layerwise utilized rank of different network-dataset pairs suggests that most networks do not fully utilize the weight-space available to them. This means that a straightforward low rank decomposition can significantly reduce the number of parameters and FLOPs. For instance, we show that ViT variants trained on ImageNet only have 20\% - 35\% \textit{mean layer utilization}, and can be decomposed to 25 to 48\% of their original size while reducing the original FLOPs by between 13 to 33\%. The resulting drop in accuracy after finetuning is less than 0.2\%. We find that self-supervised pretraining uses the available space better ($MLU =69\%)$, making it suitable for multiple downstream tasks. We also study the effect of scaling the network and  of increasing the dataset complexity.

\begin{figure*}[t]
\centering
   \begin{subfigure}{0.65\textwidth} 
   \centering
     \includegraphics[width=\textwidth]{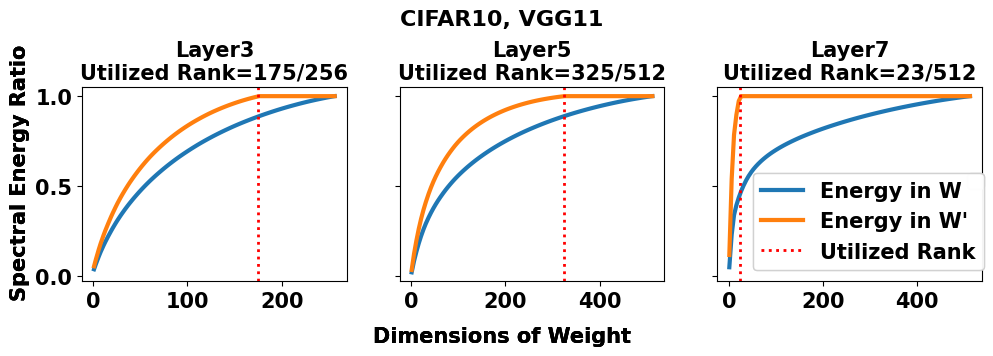}
     \vspace{-3mm}
     \caption{Spectral energy spread of W and W'. The utilized rank becomes easily identifiable when we transform W to W'}\label{fig:spread}
   \end{subfigure} \hfill
   \begin{subfigure}{0.32\textwidth} 
   \centering
     \includegraphics[width=\textwidth]{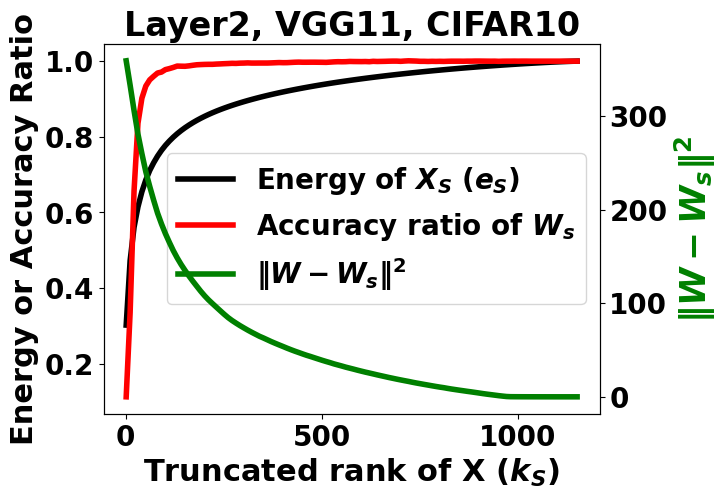}
     \vspace{-3mm}
     \caption{Projecting W onto S resulting from truncating the rank of X}\label{fig:monotonic}
   \end{subfigure}
   \vspace{-1mm}
\caption{Experiments on different layers of VGG11, CIFAR10} \label{fig:twofigs}
\vspace{-2mm}
\end{figure*}
\section{Methodology}
\label{sec:meth}

\subsection{Preliminaries: The Input and Output Subspaces}
\label{subsection3.1}
For simplicity, we consider a fully connected layer of a neural network. Let the input be to this layer be $X \in \mathbb{R}^{B \times d}$, where $B$ is the batch size and each row vector  $x \in \mathbb{R}^d$. Let $W^T \in \mathbb{R}^{d\times m}$ be the weight that maps $X$ from to $Y \in \mathbb{R}^{B \times m}$ 
. The corresponding forward pass can be written as:
\begin{equation}
\label{eq:forward_pass}
Y=XW^T
\end{equation}
For the first layer of a neural network, $X$ is real data such as images.  Similarly, the output of the layer would be dependent on the overlap between the input space and the column space of $W$, i.e. if the columns of $W$ and $X$ were orthogonal, the output would be zero. Generalizing to all layers, let $S$ be the subspace of the input to the layer, with dimension $k_S$. The orthogonal complement of this subspace, $S_\perp$ is $d-k_S$ dimensional, and contains the space of inputs or activations not occupied by the real input. We can find this subspace using SVD, shown below
\begin{equation}
    X = U_X \Sigma_X V_X^T
\end{equation}
where $\Sigma_X$ is a diagonal matrix of the $d$ singular values $\sigma_i$ and the first $k_S$ rows of $V_X^T$ represents a bases for $S$. We define the spectral energy ratio as $e_S = \frac{\sum_{0}^{k_S} \sigma_i^2}{ \sum_{0}^{d} \sigma_i^2}$ such that  we can preserve 99\% of the spectral energy e.g. $e_s = 0.99$ with $k_S$ equal to the number of singular values (squared) that contain 99\% of the total energy.  We construct the projection matrix $P_S$ that projects $X$ onto $S$, denoted by $X_S$ as:
\begin{align}
V_S := V_X[:k_S]; &\quad V_{S_\perp} := V_X[k_S:] \\
P_S = V_S^TV_S; &\quad  P_{S_\perp} = V_{S_\perp}^TV_{S_\perp} \\ 
X_S = X P_S &\quad X_{S_\perp} = X P_{S_\perp} 
\end{align}

\vspace{-2mm}
Similarly, let the subspace of the output be $T$ and the spectral energy $e_T$ correspond to the utilized rank $k_T$. Similar to equations for $S$, $V_T$ contains the bases for $T$ found from performing the SVD on $Y$ and gives the projection matrix for $P_T \in \mathbb{R}^{m\times m}$. Further details for SVD computation are provided in appendix section \ref{app:svd_deets}.


\begin{figure*}
    \centering
    \includegraphics[width=1\textwidth]{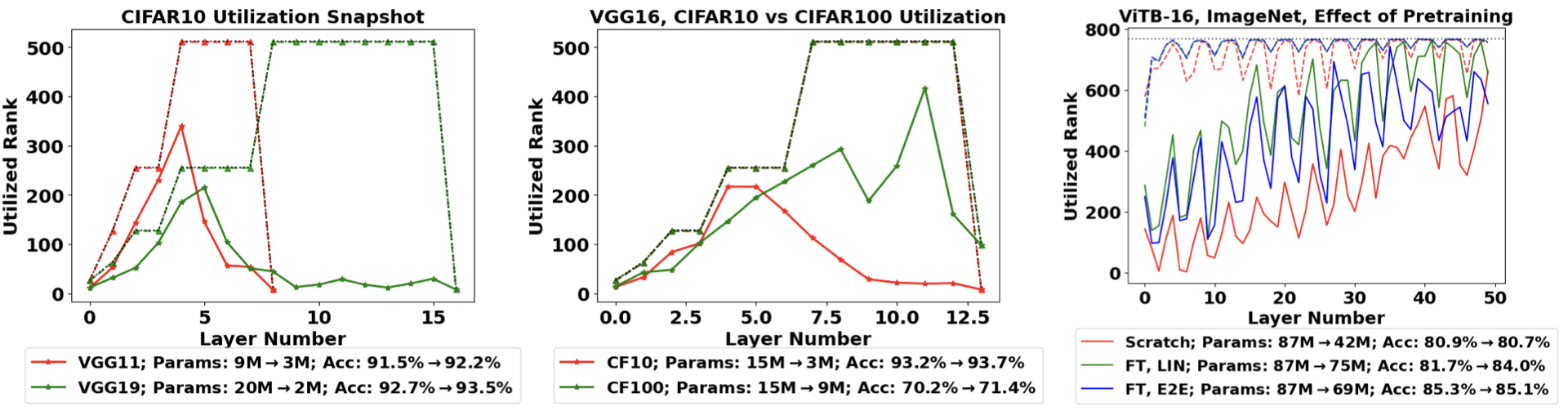}
    \vspace{-5mm}
    \caption{Utilization snapshots of different dataset-network pairs.The rank of the unaltered W is plotted for each layer in the dotted line, and the rank of the transformed W as the solid line. The brackets list the parameters and accuracy of the original and decomposed, finetuned network. FT refers to finetuning from SWAG \protect\cite{swag}, in a linear or end-to-end fashion.}
    \label{fig:snapshots}
\end{figure*}
\subsection{The Weight Transformation and the Utilized Rank}
\label{subsection3.2}

In the forward pass equation \ref{eq:forward_pass}, replacing $X$ with $X_S$ and $Y$ with $Y_T$ should result in the forward pass mapping remaining largely unaltered. This is equivalent to modifying $W$ by projecting its column space onto $S$ and its row space onto $T$, resulting in a modified $W'$ as shown below:
\begin{align}
    Y \approx Y_T &= YP_T \\
     &= XW^TP_T\\ 
    &\approx X P_S W^TP_T \\
    \implies Y &\approx XW'^T \texttt{;   where   } W':= P_SW^TP_T  
\end{align} 
We refer to the rank of $W'$ as the \textit{utilized rank} since \textbf{this transformation is data-dependent and captures the subspace overlap between the weight-space and the data-space}. In Figure \ref{fig:spread}, we show the spectral energy distribution of $W$ and $W'$ for different layers of VGG11 trained on CIFAR10 data. From the figure, we can see that \textbf{the spectral energy of \textcolor{blue}{$W$} has a wider distribution than \textcolor{orange}{$W'$}, obfuscating the true rank. Transforming $W$ into $W'$ compacts the spectral energy and allows us to identify the utilized rank more easily that naively applying the SVD directly}. For later layers, we note that the utilized rank is a small fraction of the available dimensions ($23/512$), highlighting the overparametrization of VGG architectures for CIFAR10. 
The resulting error from replacing $W$ by $W'$ in equation \ref{eq:forward_pass} can be upper-bounded by choosing appropriate dimensions for the input and output subspaces $k_S$ and $k_T$.
\begin{align}
\|E\|^2 &= \|XW^T - X (P_T W P_S)^T \|^2 \\
&\leq (1-e_T)\|Y\|^2 + (1-e_S) \|X\|^2 \|W\|^2 
\end{align}
The proof utilizes the fact that the Frobenius norm is the sum of the square of singular values (see Appendix section \ref{app:proof}).

\definecolor{ForestGreen}{rgb}{0.13, 0.55, 0.13} 
\paragraph{How to choose $k_S$ and $k_T$} The error per layer is a function of $e_S$ and $e_T$, and we use validation accuracy to inform us of the maximum $k_S$ and $k_T$ we can set before suffering a performance drop. In Figure \ref{fig:monotonic}, we vary $k_S$ for a single layer of VGG11 trained on CIFAR10 and plot the impact on $e_S$ (black), the accuracy when we replace $W$ by $W_S$ as a ratio of the original accuracy (\textcolor{red}{red}), and the norm difference between $W$ and $W_S$ (\textcolor{ForestGreen}{green}). For this layer, we note that when $k_S$ reaches $\approx 200/1024$ dimensions, the transformed $W_S$ does not result in an accuracy drop even though $W_S$ differs significantly from $W$ in norm ($\approx 150)$. When $k_S=200$ and $e_S=0.8$, retaining only 80\% of the energy was sufficient to achieve full accuracy. 
Hence, to maximize savings, we perform a binary search on $e_S$ and $e_T$ for each layer, while using validation accuracy drop as the signal to inform the stopping criterion. We call the accuracy drop tolerance for each transformation, ($S$ or $T$ projection for each layer) as $\epsilon$, and set it to 0.1 for our experiments. 
After estimating $k_S,\ k_T$ that conforms to this $\epsilon$ error for all layers, the transformed network would have an accuracy drop = $2 \times \#\text{layers} \times \epsilon\%$, which scales with the depth of the network. However, \textbf{since we largely preserve the functional mapping of each layer, we find that finetuning is able to recover the allocated drop}. When finetuning, we decompose each layer into 2 layers of reduced rank to ensure that finetuning does not increase the searched rank.


\subsection{Benefits of Studying the Utilized Ranks of Layers}

\textbf{Mean Layer Utilization:}  We describe the utilization statistic for a layer as the ratio of the rank of $W'$ to the maximum rank possible. Suppose the utilized rank of a layer with  $W \in \mathbb{R}^{m \times d}$ is $r$, then the layer utilization is $\frac{r}{\min(m,\ d)}$.  The rank of $W'$ is constrained to the rank of the product of $P_SW^TP_T$, so we can calculate the rank $r$ for a given layer as $\min(k_S,\ k_T)$. 
A utilization close to 1 implies that the learnt column space of the weight overlaps fully with the subspace of the input to the layer, whereas a utilization close to 0 implies that the spaces are orthogonal, resulting in little to no signal being passed forward. The utilized rank depends on both the network architecture and the dataset, allowing us to capture a notion of capacity that is more informative than just the number of FLOPs or parameters. We average this score over all convolutional and linear layers, and call this the MLU (mean layer utilization) score of the network. \textbf{A higher MLU reveals that the network is well utilized, while a lower MLU allows for low-rank decomposition for efficiency.}

\textbf{Savings in FLOPs and parameters:} This low dimensionality of $W$ results in a low rank decomposition that directly reduces memory and compute costs if the rank $r \leq \frac{m \times d}{m+d}$. Hence, for all layers that meet this criterion, we decompose the layer into 2 layers with weights of shapes $r \times d$ and $m\times r $, respectively. This reduces the total parameters and compute approximately by a factor of $ \frac{(m \times d)}{r (m+d)}.$  

\textbf{Utilization Snapshot:} To study the layer-specific dynamics of rank utilization, we chart the rank of the learned $W$, the utilized rank $r$, and the maximum rank possible at each layer as a utilization snapshot of a trained network. This can visualize the maximum per-layer utilization across various network and dataset combinations.  We can also utilize this to understand the effects of different pretraining and finetuning techniques.



\begin{table*}[t]
\small
\centering
\caption{Results for Utilized Rank Decomposition on ImageNet. ViT \cite{vit} and ResNet \cite{resnet, wideresnet} pretrained models from torchvision \cite{pytorch}, DeiT \cite{deit} and SWIN\cite{swin} from TIMM \cite{pytorch} *implies $\epsilon = 0.05\%, 0.1\%$ otherwise. $^\dagger$Finetuning the original DeiT models results in improved performance.}
\label{tab:imnet}
\small
\addtolength{\tabcolsep}{3.5pt}
\def\arraystretch{1.25}%
\begin{tabular}{|
>{\columncolor[HTML]{D4D4D4}}c cccccc|}
\hline
\cellcolor[HTML]{B0B3B2}Architecture &
  \cellcolor[HTML]{B0B3B2}\begin{tabular}[c]{@{}c@{}}Orig Acc\\  (\%)\end{tabular} &
  \cellcolor[HTML]{B0B3B2}\begin{tabular}[c]{@{}c@{}}Orig MLU\\  (\%)\end{tabular} &
  \cellcolor[HTML]{B0B3B2}\begin{tabular}[c]{@{}c@{}}Acc - Ours\\  (\%) ($\Delta$)\end{tabular} &
  \cellcolor[HTML]{B0B3B2}\begin{tabular}[c]{@{}c@{}}True MLU\\  (\%)\end{tabular} &
  \cellcolor[HTML]{B0B3B2}\begin{tabular}[c]{@{}c@{}}Params\\  Ratio\end{tabular} &
  \cellcolor[HTML]{B0B3B2}\begin{tabular}[c]{@{}c@{}}Flops\\  Ratio\end{tabular} \\ \hline
ViTB16           & 80.9 & 94 & 80.7 (-0.2) & 35 & 0.48 & 0.33 \\
ViTB32           & 75.7 & 94 & 75.8 (+0.1) & 34 & 0.46 & 0.33 \\
ViTL16           & 79.5 & 81 & 79.5 (+0.0) & 20 & 0.25 & 0.13 \\
ViTL32*          & 76.9 & 92 & 76.2 (-0.7) & 26 & 0.36 & 0.26 \\ \hline
DeiT - Tiny$^\dagger$      & 72.1 / 75.3 & 98 & 75.0 (-0.3) & 86 & 0.99 & 0.99 \\
DeiT - Small$^\dagger$     & 79.8 / 80.1 & 98 & 80.3 (+0.2) & 74 & 0.89 & 0.89 \\
DeiT - Base$^\dagger$      & 81.8 / 82.0 & 98 & 81.5 (-0.5) & 49 & 0.64 & 0.65 \\ \hline
SWIN - Tiny      & 81.2 & 98 & 81.3 (+0.1) & 65 & 0.86 & 0.83 \\
SWIN - Small     & 83.3 & 98 & 83.4 (+0.1) & 60 & 0.81 & 0.77 \\
SWIN- Base*      & 85.2 & 98 & 84.5 (-0.7) & 66 & 0.86 & 0.83 \\
SWIN - Large*    & 86.3 & 98 & 85.3 (-1.0) & 53 & 0.74 & 0.70 \\ \hline
ResNet34         & 73.2 & 99 & 72.2 -(1.0) & 66 & 0.77 & 0.76 \\
ResNet50         & 80.1 & 99 & 79.4 (-0.7) & 60 & 0.83 & 0.74 \\
ResNet101        & 81.5 & 99 & 80.5 (-1.0) & 47 & 0.66 & 0.59 \\
WideResNet50\_2  & 81.2 & 99 & 80.6 (-0.6) & 43 & 0.68 & 0.58 \\
WideResNet101\_2 & 82.3 & 99 & 81.7 (-0.6) & 33 & 0.51 & 0.44 \\ \hline
\end{tabular}
\end{table*}

\section{Results and Discussion}
\label{sec:results}

We perform experiments on VGG \cite{vgg}, ResNet \cite{resnet}, ViT \cite{vit}, DeiT \cite{deit}, Swin Transformer \cite{swin}, and Resnet variants \cite{wideresnet} on CIFAR10, CIFAR100 \cite{cifar}, and ImageNet \cite{imagenet}. We use pretrained ViTs and ResNets from torchvision \cite{pytorch} and DeiTs and SWIN transformers from TIMM \cite{timm}\footnote{For CIFAR, we use the architectures and hyperparameters from github.com/bearpaw/pytorch-classification}.  We use Deepspeed \cite{deepspeed} for profiling FLOPs with a batch size of 32. We define the drop per layer at $\epsilon= 0.1\%$. For ViTL-32, Swin-Base, and Swin-Large, the finetuned accuracy drop for $\epsilon=0.1\%$ was greater than 1\%, and was reduced to 0.05\%. We use SVD for calculating ranks. To rule out very small singular values arising from numerical errors, we assign the rank as the number of singular values that explain 99.99\% spectral energy. Finetuning is done with each layer decomposed into two layers of reduced rank to ensure it does not increase rank. However, when reporting final savings, we decompose only those layers where matrix decomposition would result in a reduction in parameters. Finetuning hyperparameters are in Appendix section \ref{app:hyperparams}.

\subsection{Utilization Statistics of Popular Networks}

Studying layerwise utilization can help us understand the suitability of the model for the dataset. In Figure \ref{fig:snapshots}, left, we show the layer-utilization for VGG11 and VGG19, for the same dataset CIFAR10. We see that they achieve similar layer utilization, with a peak in utilization around layers 4-6 for the same task. While the original parameters grow from 9M to 20M, the utilized parameters stay  stable around 2.5M.
In Figure \ref{fig:snapshots}, center, we evaluate the effect of increasing dataset complexity on a static architecture to illustrate higher network utilization for CIFAR100 than CIFAR10.  Not only is the utilization for CIFAR100 higher, but the utilization at higher layer numbers could indicate the usage of higher level features required to solve a more complex task.  

From Tables \ref{tab:imnet} and \ref{tab:cf10}, we note that the original models have close to $100\% MLU$, deceptively implying that all the space available for learning is well used. However,\textbf{ upon decomposition, we find that the corresponding MLUs are quite low}, dipping to 20-35\% for ViT variants on ImageNet. The fact that ViTs are too big for ImageNet has been noted previously, with the popularity of `Tiny' variants.In fact, DeiT-Tiny utilizes space quite well (99\% true MLU compared to ViTL-16's 20\%), indicating that increasing size would indeed result in a gain in accuracy. We note that DeiT networks show improved performance when training for longer. For a fair comparison, we finetune DeiT pretrained models from TIMM using the same hyperparameters as ours, and compare against the finetuned models. Both this original and finetuned accuracy for DeiT models is reported.

\begin{figure*}
    \centering
    \includegraphics[width=1\textwidth]{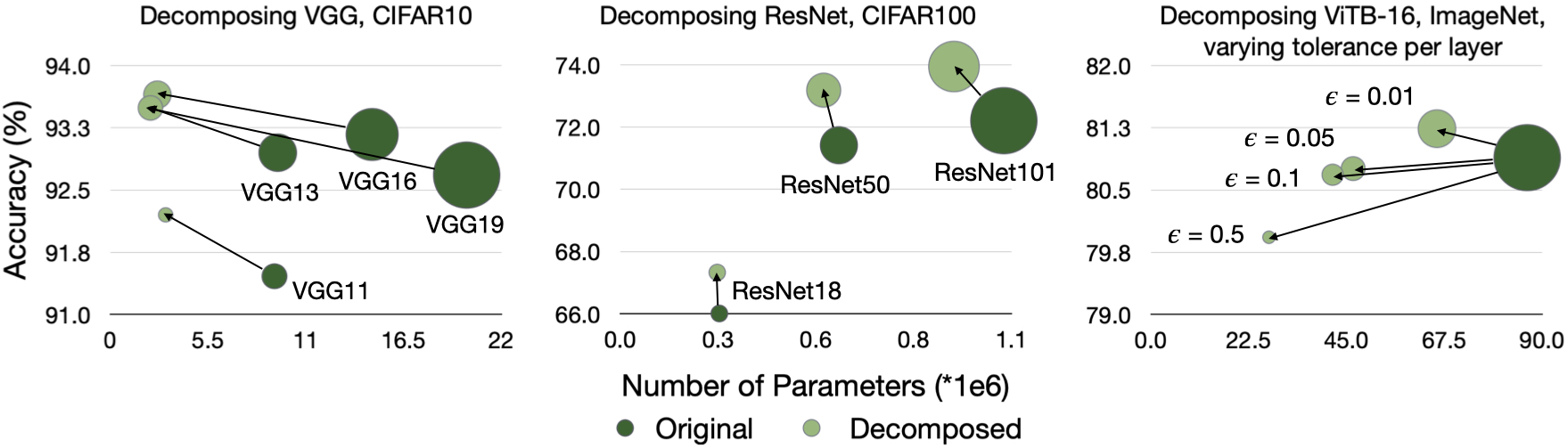}
    \vspace{-3mm}
    \caption{Visualizing the change in accuracy, number of parameters and FLOPs (size of bubble) of the decomposed, finetuned model. $\epsilon$ is the accuracy drop tolerance per layer during rank search.}
    \label{fig:bubble-graph}
    \vspace{-2mm}
\end{figure*}
\subsection{Parameter and Compute Efficiency}

In Figure \ref{fig:bubble-graph}, we study the effect of rank-decomposed and finetuned models on different architecture-dataset pairs. We plot the number of parameters against the accuracy, with the number of FLOPs represented by the sizes of the bubbles.  We see that most networks shrink and move towards the top left corner when decomposed and finetuned, implying an increase in accuracy and decrease in number of parameters and FLOPs. From tables \ref{tab:imnet} and \ref{tab:cf10}, we note that we can significantly reduce the size and FLOPs for most networks. For instance, VGG19 on CIFAR10 can be reduced to just 11\% of the original size, consuming only 38\% of the original FLOPs.  Similarly, parameters reduce to 25\% and FLOPs to 16\% on ViTL-16 for ImageNet.  On ImageNet, we see drops and increases in accuracy of less than 1\% On CIFAR, we note that finetuning accuracies never drop compared to original, sometimes increasing up to 2\% over the baseline. We attribute this potentially to an increased regularization effect from using low rank weights for small datasets. 

\subsection{Scaling network size and dataset complexity}
We show the effect of scaling a network in the same family for the same dataset in Figure \ref{fig:bubble-graph}, left, with numbers in Table \ref{tab:cf10}. We see that VGG13, VGG16, and VGG19 all converge to very similarly sized models on CIFAR10 with a very similar accuracy upon decomposition, despite being different in their original format. This indicates that a bigger network is not necessarily beneficial for CIFAR10.
However, we note that all networks report 10-20\% higher MLU when we scale up the dataset complexity, going from CIFAR10 to CIFAR100, also seen in in Figure \ref{fig:bubble-graph}, center.  This implies that the available capacity is being better utilized by larger datasets. Hence, our method serves to incorporate both the notion of capacity of the network, and its interaction with the complexity of the dataset.

\subsection{Varying the acceptable accuracy drop per layer}

We set the acceptable accuracy drop per layer, $\epsilon$, to $0.1\%$, resulting in a total accuracy drop of $0.2\% \times \#$layers. In Figure \ref{fig:bubble-graph}, we show the effect of increasing or decreasing this hyperparameter for ViTB-16  (numbers in Appendix Table \ref{tab:diffdrop}). Even when using a smaller drop of $0.01\%$ per layer, we can still reduce the network to 76\% of the parameters and 58\% of the FLOPs while gaining 0.4\% accuracy improvement, indicating that ViTB-16 is too large of a network for ImageNet. The smallest model resulting with $\epsilon=0.5\%$ consumes only $31\%$ of the parameters and $20\%$ of the original FLOPs, and shows an accuracy drop of less than 1\%. While $\epsilon$ should be tuned for every model and dataset pair, we find that $0.1\%$ and $0.05\%$ give good results across various architectures and datasets. 

\subsection{Effect of pretraining on ViTs}

In Figure \ref{fig:snapshots}, right, we evaluate the impact of weakly supervised pretraining (SWAG \cite{swag}) on layer utilization on downstream tasks. All models start close to maximum rank shown in the dotted lines. \textcolor{ForestGreen}{[FT-LIN]} refers to the network that was frozen after pretraining with only a linear head finetuned on ImageNet. The frozen weights learned from self supervised pretrained utilize the available space to the highest extent  ($MLU=69\%$), reflecting its suitability for downstream tasks. 
The model finetuned, end-to-end on ImageNet \textcolor{blue}{[FT-E2E]} shows a drop in layer-utilization, especially at later layers, since it is altered for the classification task. Training a model from random initialization \textcolor{red}{[scratch]} yields a bespoke model for ImageNet and shows lower layer utilization ($MLU=35\%$). The increase in accuracy for the LIN-FT network using our method is an unfair comparison, since we finetune end-to-end after finding the rank.




\section{Conclusion }
In this work, we proposed the \textit{mean layer utilization}, a simple data-dependent metric for determining how efficiently a neural network learns a particular dataset.  We do this by creating projection matrices for each layer to transform the learned weights onto a compact subspace dictated by the input and output activations with a controllable error that is upper bounded by the spectral energy of the input and output subspaces $e_S$ and $e_T$.  This compact representation reveals what we call the \textit{utilized rank} of a matrix, which serves as a notion of capacity that includes both the network architecture and the dataset. 
Lastly, decomposing the layers onto these data-dependent subspaces naturally lend themselves to a simple weight matrix factorization which can easily be applied to various popular network architectures such as ViTs and ResNets achieving significant parameter reduction without compromising on downstream task performance. 

\clearpage

\label{sec:References}
\nocite{*}

\bibliographystyle{abbrvnat}
\bibliography{main}

\begin{thebibliography}{61}
\providecommand{\natexlab}[1]{#1}
\providecommand{\url}[1]{\texttt{#1}}
\expandafter\ifx\csname urlstyle\endcsname\relax
  \providecommand{\doi}[1]{doi: #1}\else
  \providecommand{\doi}{doi: \begingroup \urlstyle{rm}\Url}\fi

\bibitem[Aghajanyan et~al.(2020)Aghajanyan, Zettlemoyer, and Gupta]{intrinsicdimensionfinetune}
A.~Aghajanyan, L.~Zettlemoyer, and S.~Gupta.
\newblock Intrinsic dimensionality explains the effectiveness of language model fine-tuning, 2020.

\bibitem[Ashkboos et~al.(2024)Ashkboos, Croci, do~Nascimento, Hoefler, and Hensman]{slicegpt}
S.~Ashkboos, M.~L. Croci, M.~G. do~Nascimento, T.~Hoefler, and J.~Hensman.
\newblock Slicegpt: Compress large language models by deleting rows and columns, 2024.

\bibitem[Brown et~al.(2023)Brown, Williamson, Anderson, and Lawrence]{transformer_kd}
N.~Brown, A.~Williamson, T.~Anderson, and L.~Lawrence.
\newblock Efficient transformer knowledge distillation: A performance review, 2023.

\bibitem[Cai et~al.(2008)Cai, Candes, and Shen]{Cai2008}
J.-F. Cai, E.~J. Candes, and Z.~Shen.
\newblock A singular value thresholding algorithm for matrix completion, 2008.

\bibitem[Deng et~al.(2009)Deng, Dong, Socher, Li, Li, and Fei-Fei]{imagenet}
J.~Deng, W.~Dong, R.~Socher, L.-J. Li, K.~Li, and L.~Fei-Fei.
\newblock Imagenet: A large-scale hierarchical image database.
\newblock pages 248--255, 2009.
\newblock \doi{10.1109/CVPR.2009.5206848}.

\bibitem[Dosovitskiy et~al.(2021)Dosovitskiy, Beyer, Kolesnikov, Weissenborn, Zhai, Unterthiner, Dehghani, Minderer, Heigold, Gelly, Uszkoreit, and Houlsby]{vit}
A.~Dosovitskiy, L.~Beyer, A.~Kolesnikov, D.~Weissenborn, X.~Zhai, T.~Unterthiner, M.~Dehghani, M.~Minderer, G.~Heigold, S.~Gelly, J.~Uszkoreit, and N.~Houlsby.
\newblock An image is worth 16x16 words: Transformers for image recognition at scale, 2021.

\bibitem[Feng et~al.(2022)Feng, Zheng, Huang, Zhao, Jordan, and Zha]{rankdiminish}
R.~Feng, K.~Zheng, Y.~Huang, D.~Zhao, M.~Jordan, and Z.-J. Zha.
\newblock Rank diminishing in deep neural networks, 2022.

\bibitem[Frankle and Carbin(2018)]{lth}
J.~Frankle and M.~Carbin.
\newblock The lottery ticket hypothesis: Finding sparse, trainable neural networks.
\newblock \emph{arXiv preprint arXiv:1803.03635}, 2018.

\bibitem[Frantar and Alistarh(2023)]{sparsegpt}
E.~Frantar and D.~Alistarh.
\newblock Sparsegpt: Massive language models can be accurately pruned in one-shot, 2023.

\bibitem[Frantar et~al.(2023)Frantar, Ashkboos, Hoefler, and Alistarh]{gptq}
E.~Frantar, S.~Ashkboos, T.~Hoefler, and D.~Alistarh.
\newblock Gptq: Accurate post-training quantization for generative pre-trained transformers, 2023.

\bibitem[Fukushima(1969)]{relu-bio}
K.~Fukushima.
\newblock Visual feature extraction by a multilayered network of analog threshold elements.
\newblock \emph{IEEE Transactions on Systems Science and Cybernetics}, 5:\penalty0 322--333, 1969.
\newblock \doi{10.1109/TSSC.1969.300225}.

\bibitem[Garg et~al.(2020)Garg, Panda, and Roy]{Garg2020}
I.~Garg, P.~Panda, and K.~Roy.
\newblock A low effort approach to structured cnn design using pca.
\newblock \emph{IEEE Access}, 8:\penalty0 1347--1360, 2020.
\newblock ISSN 2169-3536.
\newblock \doi{10.1109/access.2019.2961960}.
\newblock URL \url{http://dx.doi.org/10.1109/ACCESS.2019.2961960}.

\bibitem[Glorot and Bengio(2010)]{xavier_init}
X.~Glorot and Y.~Bengio.
\newblock Understanding the difficulty of training deep feedforward neural networks.
\newblock In Y.~W. Teh and M.~Titterington, editors, \emph{Proceedings of the Thirteenth International Conference on Artificial Intelligence and Statistics}, volume~9 of \emph{Proceedings of Machine Learning Research}, pages 249--256, Chia Laguna Resort, Sardinia, Italy, 13--15 May 2010. PMLR.
\newblock URL \url{https://proceedings.mlr.press/v9/glorot10a.html}.

\bibitem[Grasedyck et~al.(2013)Grasedyck, Kressner, and Tobler]{lowranksurvey}
L.~Grasedyck, D.~Kressner, and C.~Tobler.
\newblock A literature survey of low-rank tensor approximation techniques, 2013.

\bibitem[He et~al.(2015)He, Zhang, Ren, and Sun]{resnet}
K.~He, X.~Zhang, S.~Ren, and J.~Sun.
\newblock Deep residual learning for image recognition, 2015.

\bibitem[Hinton et~al.(2015)Hinton, Vinyals, and Dean]{kd}
G.~Hinton, O.~Vinyals, and J.~Dean.
\newblock Distilling the knowledge in a neural network, 2015.

\bibitem[Hoffer et~al.(2018)Hoffer, Banner, Golan, and Soudry]{wd1}
E.~Hoffer, R.~Banner, I.~Golan, and D.~Soudry.
\newblock Norm matters: efficient and accurate normalization schemes in deep networks.
\newblock \emph{Advances in Neural Information Processing Systems}, 31, 2018.

\bibitem[Hu et~al.(2021)Hu, Shen, Wallis, Allen-Zhu, Li, Wang, Wang, and Chen]{lora}
E.~J. Hu, Y.~Shen, P.~Wallis, Z.~Allen-Zhu, Y.~Li, S.~Wang, L.~Wang, and W.~Chen.
\newblock Lora: Low-rank adaptation of large language models, 2021.

\bibitem[Huh et~al.(2023)Huh, Mobahi, Zhang, Cheung, Agrawal, and Isola]{lowrankbias}
M.~Huh, H.~Mobahi, R.~Zhang, B.~Cheung, P.~Agrawal, and P.~Isola.
\newblock The low-rank simplicity bias in deep networks, 2023.

\bibitem[Idelbayev and Carreira-Perpinan(2020)]{ranklearning}
Y.~Idelbayev and M.~A. Carreira-Perpinan.
\newblock Low-rank compression of neural nets: Learning the rank of each layer.
\newblock pages 8046--8056, 2020.
\newblock \doi{10.1109/CVPR42600.2020.00807}.

\bibitem[Jaderberg et~al.(2014)Jaderberg, Vedaldi, and Zisserman]{lowrankexpansion}
M.~Jaderberg, A.~Vedaldi, and A.~Zisserman.
\newblock Speeding up convolutional neural networks with low rank expansions, 2014.

\bibitem[Johnson and Lindenstrauss(1984)]{jl}
W.~Johnson and J.~Lindenstrauss.
\newblock Extensions of lipschitz maps into a hilbert space.
\newblock \emph{Contemporary Mathematics}, 26:\penalty0 189--206, 01 1984.
\newblock \doi{10.1090/conm/026/737400}.

\bibitem[Kamalakara et~al.(2022)Kamalakara, Locatelli, Venkitesh, Ba, Gal, and Gomez]{lowranktrain}
S.~R. Kamalakara, A.~Locatelli, B.~Venkitesh, J.~Ba, Y.~Gal, and A.~N. Gomez.
\newblock Exploring low rank training of deep neural networks, 2022.

\bibitem[Khodak et~al.(2021)Khodak, Tenenholtz, Mackey, and Fusi]{lowrankinit}
M.~Khodak, N.~Tenenholtz, L.~Mackey, and N.~Fusi.
\newblock Initialization and regularization of factorized neural layers.
\newblock \emph{arXiv preprint arXiv:2105.01029}, 2021.

\bibitem[Khodak et~al.(2022)Khodak, Tenenholtz, Mackey, and Fusi]{factornn}
M.~Khodak, N.~Tenenholtz, L.~Mackey, and N.~Fusi.
\newblock Initialization and regularization of factorized neural layers, 2022.

\bibitem[Kim et~al.(2019{\natexlab{a}})Kim, Khan, and Kyung]{kim2019efficient}
H.~Kim, M.~U.~K. Khan, and C.-M. Kyung.
\newblock Efficient neural network compression, 2019{\natexlab{a}}.

\bibitem[Kim et~al.(2019{\natexlab{b}})Kim, Khan, and Kyung]{rankpca}
H.~Kim, M.~U.~K. Khan, and C.-M. Kyung.
\newblock Efficient neural network compression.
\newblock In \emph{Proceedings of the IEEE/CVF Conference on Computer Vision and Pattern Recognition (CVPR)}, June 2019{\natexlab{b}}.

\bibitem[Krizhevsky and Hinton(2009)]{cifar}
A.~Krizhevsky and G.~Hinton.
\newblock Learning multiple layers of features from tiny images.
\newblock Technical Report~0, University of Toronto, Toronto, Ontario, 2009.
\newblock URL \url{https://www.cs.toronto.edu/~kriz/learning-features-2009-TR.pdf}.

\bibitem[Li et~al.(2018)Li, Farkhoor, Liu, and Yosinski]{intrinsicdimensionlandscape}
C.~Li, H.~Farkhoor, R.~Liu, and J.~Yosinski.
\newblock Measuring the intrinsic dimension of objective landscapes, 2018.

\bibitem[Li et~al.(2016)Li, Liang, and Risteski]{lowrankapprox}
Y.~Li, Y.~Liang, and A.~Risteski.
\newblock Recovery guarantee of weighted low-rank approximation via alternating minimization, 2016.

\bibitem[Liebenwein et~al.(2021)Liebenwein, Maalouf, Gal, Feldman, and Rus]{optimaldecompose}
L.~Liebenwein, A.~Maalouf, O.~Gal, D.~Feldman, and D.~Rus.
\newblock Compressing neural networks: Towards determining the optimal layer-wise decomposition, 2021.

\bibitem[Liu et~al.(2021)Liu, Lin, Cao, Hu, Wei, Zhang, Lin, and Guo]{swin}
Z.~Liu, Y.~Lin, Y.~Cao, H.~Hu, Y.~Wei, Z.~Zhang, S.~Lin, and B.~Guo.
\newblock Swin transformer: Hierarchical vision transformer using shifted windows, 2021.

\bibitem[Mahabadi et~al.(2021)Mahabadi, Henderson, and Ruder]{compacter}
R.~K. Mahabadi, J.~Henderson, and S.~Ruder.
\newblock Compacter: Efficient low-rank hypercomplex adapter layers, 2021.

\bibitem[Mangrulkar et~al.(2022)Mangrulkar, Gugger, Debut, Belkada, Paul, and Bossan]{peft}
S.~Mangrulkar, S.~Gugger, L.~Debut, Y.~Belkada, S.~Paul, and B.~Bossan.
\newblock Peft: State-of-the-art parameter-efficient fine-tuning methods, 2022.

\bibitem[Nair and Hinton(2010)]{relu}
V.~Nair and G.~E. Hinton.
\newblock Rectified linear units improve restricted boltzmann machines.
\newblock pages 807--814. Omnipress, 2010.
\newblock ISBN 9781605589077.

\bibitem[Noach and Goldberg(2020)]{langcompress}
M.~B. Noach and Y.~Goldberg.
\newblock Compressing pre-trained language models by matrix decomposition.
\newblock pages 884--889. Association for Computational Linguistics, 12 2020.
\newblock URL \url{https://aclanthology.org/2020.aacl-main.88}.

\bibitem[Oymak et~al.(2019)Oymak, Fabian, Li, and Soltanolkotabi]{genguarantee}
S.~Oymak, Z.~Fabian, M.~Li, and M.~Soltanolkotabi.
\newblock Generalization guarantees for neural networks via harnessing the low-rank structure of the jacobian.
\newblock \emph{arXiv preprint arXiv:1906.05392}, 2019.

\bibitem[Paszke et~al.(2019)Paszke, Gross, Massa, Lerer, Bradbury, Chanan, Killeen, Lin, Gimelshein, Antiga, Desmaison, Köpf, Yang, DeVito, Raison, Tejani, Chilamkurthy, Steiner, Fang, Bai, and Chintala]{pytorch}
A.~Paszke, S.~Gross, F.~Massa, A.~Lerer, J.~Bradbury, G.~Chanan, T.~Killeen, Z.~Lin, N.~Gimelshein, L.~Antiga, A.~Desmaison, A.~Köpf, E.~Yang, Z.~DeVito, M.~Raison, A.~Tejani, S.~Chilamkurthy, B.~Steiner, L.~Fang, J.~Bai, and S.~Chintala.
\newblock Pytorch: An imperative style, high-performance deep learning library, 2019.

\bibitem[Rasley et~al.(2020)Rasley, Rajbhandari, Ruwase, and He]{deepspeed}
J.~Rasley, S.~Rajbhandari, O.~Ruwase, and Y.~He.
\newblock Deepspeed: System optimizations enable training deep learning models with over 100 billion parameters.
\newblock In \emph{Proceedings of the 26th ACM SIGKDD International Conference on Knowledge Discovery \& Data Mining}, KDD '20, page 3505–3506, New York, NY, USA, 2020. Association for Computing Machinery.
\newblock ISBN 9781450379984.
\newblock \doi{10.1145/3394486.3406703}.
\newblock URL \url{https://doi.org/10.1145/3394486.3406703}.

\bibitem[Roy and Vetterli(2007)]{original_eff_rank}
O.~Roy and M.~Vetterli.
\newblock The effective rank: A measure of effective dimensionality.
\newblock In \emph{2007 15th European Signal Processing Conference}, pages 606--610, 2007.

\bibitem[Sainath et~al.(2013)Sainath, Kingsbury, Sindhwani, Arisoy, and Ramabhadran]{Sainath2013}
T.~N. Sainath, B.~Kingsbury, V.~Sindhwani, E.~Arisoy, and B.~Ramabhadran.
\newblock Low-rank matrix factorization for deep neural network training with high-dimensional output targets.
\newblock pages 6655--6659, 2013.
\newblock \doi{10.1109/ICASSP.2013.6638949}.

\bibitem[Schotth{\"o}fer et~al.(2022)Schotth{\"o}fer, Zangrando, Kusch, Ceruti, and Tudisco]{lowranklth}
S.~Schotth{\"o}fer, E.~Zangrando, J.~Kusch, G.~Ceruti, and F.~Tudisco.
\newblock Low-rank lottery tickets: finding efficient low-rank neural networks via matrix differential equations.
\newblock \emph{Advances in Neural Information Processing Systems}, 35:\penalty0 20051--20063, 2022.

\bibitem[Sharma et~al.(2024)Sharma, Ash, and Misra]{laser}
P.~Sharma, J.~T. Ash, and D.~Misra.
\newblock The truth is in there: Improving reasoning in language models with layer-selective rank reduction.
\newblock In \emph{The Twelfth International Conference on Learning Representations}, 2024.
\newblock URL \url{https://openreview.net/forum?id=ozX92bu8VA}.

\bibitem[Simonyan and Zisserman(2015)]{vgg}
K.~Simonyan and A.~Zisserman.
\newblock Very deep convolutional networks for large-scale image recognition, 2015.

\bibitem[Singh et~al.(2022)Singh, Gustafson, Adcock, de~Freitas~Reis, Gedik, Kosaraju, Mahajan, Girshick, Dollár, and van~der Maaten]{swag}
M.~Singh, L.~Gustafson, A.~Adcock, V.~de~Freitas~Reis, B.~Gedik, R.~P. Kosaraju, D.~Mahajan, R.~Girshick, P.~Dollár, and L.~van~der Maaten.
\newblock Revisiting weakly supervised pre-training of visual perception models, 2022.

\bibitem[Suau et~al.(2019)Suau, Zappella, and Apostoloff]{luca}
X.~Suau, L.~Zappella, and N.~Apostoloff.
\newblock Filter distillation for network compression, 2019.

\bibitem[Sui et~al.(2024)Sui, Yin, Gong, Xiao, Phan, and Yuan]{elrt}
Y.~Sui, M.~Yin, Y.~Gong, J.~Xiao, H.~Phan, and B.~Yuan.
\newblock Elrt: Efficient low-rank training for compact convolutional neural networks, 2024.

\bibitem[Touvron et~al.(2021)Touvron, Cord, Douze, Massa, Sablayrolles, and Jégou]{deit}
H.~Touvron, M.~Cord, M.~Douze, F.~Massa, A.~Sablayrolles, and H.~Jégou.
\newblock Training data-efficient image transformers \& distillation through attention, 2021.

\bibitem[Tukan et~al.(2020)Tukan, Maalouf, Weksler, and Feldman]{byesvd}
M.~Tukan, A.~Maalouf, M.~Weksler, and D.~Feldman.
\newblock Compressed deep networks: Goodbye svd, hello robust low-rank approximation, 2020.

\bibitem[Vaswani et~al.(2023)Vaswani, Shazeer, Parmar, Uszkoreit, Jones, Gomez, Kaiser, and Polosukhin]{attention}
A.~Vaswani, N.~Shazeer, N.~Parmar, J.~Uszkoreit, L.~Jones, A.~N. Gomez, L.~Kaiser, and I.~Polosukhin.
\newblock Attention is all you need, 2023.

\bibitem[Wang et~al.(2020)Wang, Li, Khabsa, Fang, and Ma]{linformer}
S.~Wang, B.~Z. Li, M.~Khabsa, H.~Fang, and H.~Ma.
\newblock Linformer: Self-attention with linear complexity, 2020.

\bibitem[Wen et~al.(2017)Wen, Xu, Wu, Wang, Chen, and Li]{wen2017coordinating}
W.~Wen, C.~Xu, C.~Wu, Y.~Wang, Y.~Chen, and H.~Li.
\newblock Coordinating filters for faster deep neural networks, 2017.

\bibitem[Wightman(2019)]{timm}
R.~Wightman.
\newblock Pytorch image models, 2019.

\bibitem[Wolf et~al.(2020)Wolf, Debut, Sanh, Chaumond, Delangue, Moi, Cistac, Rault, Louf, Funtowicz, Davison, Shleifer, von Platen, Ma, Jernite, Plu, Xu, Scao, Gugger, Drame, Lhoest, and Rush]{huggingface}
T.~Wolf, L.~Debut, V.~Sanh, J.~Chaumond, C.~Delangue, A.~Moi, P.~Cistac, T.~Rault, R.~Louf, M.~Funtowicz, J.~Davison, S.~Shleifer, P.~von Platen, C.~Ma, Y.~Jernite, J.~Plu, C.~Xu, T.~L. Scao, S.~Gugger, M.~Drame, Q.~Lhoest, and A.~M. Rush.
\newblock Huggingface's transformers: State-of-the-art natural language processing, 2020.

\bibitem[Yaguchi et~al.(2021)Yaguchi, Suzuki, Nitta, Sakata, and Tanizawa]{decompsoenet}
A.~Yaguchi, T.~Suzuki, S.~Nitta, Y.~Sakata, and A.~Tanizawa.
\newblock Decomposable-net: Scalable low-rank compression for neural networks.
\newblock International Joint Conferences on Artificial Intelligence Organization, 8 2021.
\newblock \doi{10.24963/ijcai.2021/447}.
\newblock URL \url{http://dx.doi.org/10.24963/ijcai.2021/447}.

\bibitem[Yang(2017)]{bearpaw}
W.~Yang.
\newblock pytorch-classification.
\newblock \url{https://https://github.com/bearpaw/pytorch-classification}, 2017.
\newblock Accessed: 2023-06-01.

\bibitem[Zagoruyko and Komodakis(2017)]{wideresnet}
S.~Zagoruyko and N.~Komodakis.
\newblock Wide residual networks, 2017.

\bibitem[Zangrando et~al.(2024)Zangrando, Deidda, Brugiapaglia, Guglielmi, and Tudisco]{rankcollapse}
E.~Zangrando, P.~Deidda, S.~Brugiapaglia, N.~Guglielmi, and F.~Tudisco.
\newblock Neural rank collapse: Weight decay and small within-class variability yield low-rank bias, 2024.

\bibitem[Zhang et~al.(2019)Zhang, Wang, Xu, and Grosse]{wd2}
G.~Zhang, C.~Wang, B.~Xu, and R.~Grosse.
\newblock Three mechanisms of weight decay regularization.
\newblock In \emph{International Conference on Learning Representations}, 2019.
\newblock URL \url{https://openreview.net/forum?id=B1lz-3Rct7}.

\bibitem[Zhang et~al.(2018)Zhang, Cisse, Dauphin, and Lopez-Paz]{mixup}
H.~Zhang, M.~Cisse, Y.~N. Dauphin, and D.~Lopez-Paz.
\newblock mixup: Beyond empirical risk minimization, 2018.

\bibitem[Zhang et~al.(2014)Zhang, Chuangsuwanich, and Glass]{Zhang2014}
Y.~Zhang, E.~Chuangsuwanich, and J.~Glass.
\newblock Extracting deep neural network bottleneck features using low-rank matrix factorization.
\newblock pages 185--189, 2014.
\newblock \doi{10.1109/ICASSP.2014.6853583}.

\end{thebibliography}

\clearpage

\onecolumn

\section{Appendix}
\label{sec:appendix}

\subsection{Upper-bounding the error from transforming $W$ to $W'$}
\label{app:proof}

We note that the projection matrices are symmetric since $P_S^T = (V_S^T V_S)^T = P_S$. We use these to express the error from transforming $W$ to $W'$ in terms of the perpendicular spaces.

\begin{align}
E &= XW^T - X W'^T \\
&= XW^T - X (P_T W P_S)^T \\
 &= XW^T - (X P_S) W^T P_T \\
 &= XW^T - X_S W^T P_T \\
  &= XW^T - ( X - X_{S_\perp}) W^T P_T \\
  &= (XW^T - XW^TP_T) + X_{S_\perp} W^T P_T \\
  &= (Y - Y_T) + X_{S_\perp} (P_T W)^T \\
  &= Y_{T_\perp} + X_{S_\perp}  W_T^T \label{perp-eqns-end}
\end{align}


Since the Frobenius norm of a matrix, squared, is the sum of its singular values, squared, our definition of $S, T$ implies the following relations:
\begin{align}
X &= X_S + X_{S_\perp}; & Y &= Y_T + Y_{T_\perp}; \\ 
\| X_S \|^2 &= e_S \|X \|^2; & \| Y_T \|^2 &= e_T \|Y \|^2; \\
\|X_{S_\perp} \|^2 &= (1-e_S) \| X\|^2; & \| Y_{T_\perp} \|^2 &= (1-e_T) \|Y \|^2
\end{align}

where all norms refer to Frobenius norm. Additionally, we know that $\| A + B\|_F^2 =  Tr\left((A+B)^T(A+B)\right) = \| A \|_F^2 + \|B \|_F^2 + 2Tr(A^TB) $. Since trace is invariant to cyclic permutation and transpose), we have $Tr (A^TB) = Tr(AB^T) $. Putting all this together, we can upper bound the error in equation \ref{perp-eqns-end} as follows.

\begin{align}
\|E\|^2 &= \| Y_{T_\perp} \|^2 +  \|X_{S_\perp} W_T^T\|^2 + 2Tr(Y_{T_\perp} W_T  X_{S_\perp}^T)  \\
  &= \| Y_{T_\perp} \|^2 +  \|X_{S_\perp} W_T^T\|^2 + 2Tr(Y P_{T_\perp} P_T W  X_{S_\perp}^T) \\
  &= \| Y_{T_\perp} \|^2 + \|X_{S_\perp} W_T^T\|^2 + 2Tr(Y (P_{T_\perp} P_T)   X_{S_\perp}^T) \\
  &= \| Y_{T_\perp} \|^2 + \|X_{S_\perp} W_T^T\|^2 + 0 \label{eq:trace}\\
  &= \| Y_{T_\perp} \|^2 + \|XP_{S_\perp}  W_T^T\|^2 \\
&\leq \| Y_{T_\perp} \|^2 + \|X_{S_\perp} \|^2 \|W_T^T \|^2  \\
&\leq \| Y_{T_\perp} \|^2 + \|X_{S_\perp} \|^2 \|W^T \|^2  \label{ineqW} \\
&= (1-e_T)\|Y\|^2 + (1-e_S) \|X\|^2 \|W\|^2 
\end{align}

The trace in equation \ref{eq:trace} reduces to zero since we multiply two matrices in orthogonal spaces, resulting in zero. The last inequality in equation \ref{ineqW} arises from applying triangle inequality on $W$.
\begin{align}
W &= W_T + W_{T_\perp} \\
\| W \|^2  &= \| W_T \|^2 + \|W_{T_\perp}\|^2 + 2 Tr(W_T W_{T_\perp}) \\
\| W \|^2  &= \| W_T \|^2 + \|W_{T_\perp}\|^2 + 0 \\
\| W \|^2  &\geq \|W_T\|^2
\end{align}

\subsection{Details of SVD to find bases}
\label{app:svd_deets}

For computational ease, we perform the SVD of $X^TX$, which directly gives us the bases and the square of the singular values. This only require storing the sum of $X^T X$ at each layer, which can be parallelized over multiple batches of forward passes. We do not need to store the outputs of a layer, since we can find $T^T T$ from pre and post multiplying the saved $X^T X$ with $W^T$ and $W$ respectively, and then performing SVD on this smaller matrix. For CIFAR datasets, we use the entire training dataset to perform PCA, and for ImageNet, we choose 200 samples per class, resulting in 20,000 samples. Because this computation is parallelizable across batches and requires only forward passes, the cost of finding bases and ranks of a space is negligible. Note The same analysis will hold for bias/convolutional layer with the input being the flattened patches convolved into the filters. The addition of bias back into the analysis also does not alter the subspaces under consideration, since we only look at each layer's input and output in isolation from all other layers.

\subsection{Computational Overhead of Binary Search for Rank}
\label{app:comp_overhead}
There are three main overheads: performing SVD at each layer, weight transformation and binary search on dimensions. We perform highly parallelized SVD on the entire training dataset of CIFAR, or 20,000 samples for ImageNet, and performing SVD for all layers takes lesser time than a training epoch in most cases. Each choice of $e_S$ and $e_T$ results in an analytical weight transformation from just 2 matrix multiplications, and we only need to perform a validation pass for each level of binary search to find the direction of binary search. There are a few hyperparamters that can be optimized to speed this up, such as size of data to perform SVD on, maximum levels of binary search, and conditions to quit search on, such as acceptable accuracy drop and limiting the change in dimensions between consecutive iterations. 

The most expensive part of our computation is the validation accuracy checks for binary search for rank. Let the weight matrix at a layer be $m \times d$ dimensional, with $L$ layers in the network. For the first projection on S, we perform SVD on a $d \times d$ matrix, and a binary search on the resulting $d$ singular values. Each level of binary search performs one projection to get $W'$ and one validation accuracy check. This means that we have O(log d) validation accuracy check. Similarly for the output, we have O(log m) accuracy checks, bringing the total to $L\times O(m\times d)$ accuracy checks.  
For ViTB-16, the largest layers are $768 \times 3072$, and there are approximately 50 linear layers. This means that we perform $\sim 1000$ valiation accuracy checks for this network. It took us 7.5 hours on a machine with 8 A100 GPUs to calculate the utilized rank of all layers via this binary search. 

\subsection{Hyperparameters for Finetuning}
\label{app:hyperparams}
After performing binary search on all layers of the network, we decompose each linear and convolutional layer into two consecutive layers (without non-linearity in between) so that we can finetune while preserving the searched rank. We initialize the two layers to the left and right matrices arising from SVD on the weight (with either one appropriately scaled by the singular values). We then perform a grid search on the following parameters for finetuning: learning rate, weight decay and EMA (exponential moving average) decay. When we use EMA, we start averaging the model for EMA from the beginning of finetuning. For all other hyperparameters, we used the same as the base repository that we took the model from.

\begin{table*}[h]
\centering
\small
\caption{Hyperparameters for finetuning the decomposed models.}
\addtolength{\tabcolsep}{0pt}
\def\arraystretch{1.3}%
\begin{tabular}{|
>{\columncolor[HTML]{D4D4D4}}c ccc|}
\hline
\cellcolor[HTML]{B0B3B2} &
  \cellcolor[HTML]{B0B3B2}\begin{tabular}[c]{@{}c@{}}ViTs, SWINs, DeiTs\\  on ImageNet\end{tabular} &
  \cellcolor[HTML]{B0B3B2}\begin{tabular}[c]{@{}c@{}}ResNets \\ on ImageNet\end{tabular} &
  \cellcolor[HTML]{B0B3B2}\begin{tabular}[c]{@{}c@{}}All architectures \\ on CIFAR\end{tabular} \\ \hline
Optimizer &
  adamW &
  \begin{tabular}[c]{@{}c@{}}SGD, LR decayed \\ by 0.1 at 30 and 60\end{tabular} &
  \begin{tabular}[c]{@{}c@{}}SGD, LR decayed \\ by 0.1 at 50, 100, 130\end{tabular} \\
Batch Size    & 512                   & 32                     & 32                   \\
Epochs        & 300                   & 90                     & 200                  \\
Learning Rate & {[}0.0003,0.0001{]}   & {[}0.1, 0.01, 0.001{]} & {[}0.1,0.01,0.001{]} \\
Weight Decay  & {[}0.3,0{]}           & {[}0,0.0001{]}         & {[}50, 100, 130{]}   \\
EMA Decay     & {[}0.85, 0.9, 0.95{]} & NA                     & NA                   \\ \hline
\end{tabular}
\end{table*}

\subsection{CIFAR Results}
\label{app:tab:cf10}
Here we present the numbers corresponding to the graphs in Figures \ref{fig:bubble-graph} for CIFAR10 and CIFAR100 on VGG and ResNet architecutre variants. All results correspond to networks decomposed and finetuned to respect the rank found from binary search.
\begin{table*}[h]
\caption{Results for Utilized Rank Decomposition on CIFAR dataset for different architectures.}
\label{tab:cf10}
\small
\addtolength{\tabcolsep}{1.5pt}
\def\arraystretch{1.3}%
\begin{tabular}{|
>{\columncolor[HTML]{B0B3B2}}c 
>{\columncolor[HTML]{D4D4D4}}c cccccc|}
\hline
\cellcolor[HTML]{B0B3B2} &
  \cellcolor[HTML]{B0B3B2} &
  \cellcolor[HTML]{B0B3B2}\begin{tabular}[c]{@{}c@{}}Orig Acc\\  (\%)\end{tabular} &
  \cellcolor[HTML]{B0B3B2}\begin{tabular}[c]{@{}c@{}}Orig MLU\\  (\%)\end{tabular} &
  \cellcolor[HTML]{B0B3B2}\begin{tabular}[c]{@{}c@{}}Acc - Ours\\  (\%) ($\Delta$)\end{tabular} &
  \cellcolor[HTML]{B0B3B2}\begin{tabular}[c]{@{}c@{}}True MLU\\  (\%)\end{tabular} &
  \cellcolor[HTML]{B0B3B2}\begin{tabular}[c]{@{}c@{}}Params\\  Ratio\end{tabular} &
  \cellcolor[HTML]{B0B3B2}\begin{tabular}[c]{@{}c@{}}Flops\\  Ratio\end{tabular} \\ \hline
\cellcolor[HTML]{B0B3B2}                           & VGG11     & 91.5 & 98 & 92.5 (+1.0) & 47 & 0.34 & 0.60 \\
\cellcolor[HTML]{B0B3B2}                           & VGG13     & 92.9 & 98 & 93.5 (+0.6) & 47 & 0.24 & 0.60 \\
\cellcolor[HTML]{B0B3B2}                           & VGG16     & 93.2 & 99 & 93.6 (+0.4) & 44 & 0.18 & 0.54 \\
\cellcolor[HTML]{B0B3B2}                           & VGG19     & 92.7 & 99 & 93.6 (+0.9) & 32 & 0.11 & 0.38 \\
\cellcolor[HTML]{B0B3B2}                           & ResNet18  & 90.9 & 95 & 91.4 (+0.5) & 80 & 0.86 & 0.92 \\
\cellcolor[HTML]{B0B3B2}                           & ResNet50  & 92.8 & 96 & 93.1 (+0.3) & 64 & 0.78 & 0.78 \\
\multirow{-7}{*}{\cellcolor[HTML]{B0B3B2}\rotatebox[origin=c]{90}{CIFAR10}}  & ResNet101 & 93.2 & 96 & 94.1 (+0.9) & 51 & 0.73 & 0.64 \\ \hline
\cellcolor[HTML]{B0B3B2}                           & VGG11     & 66.9 & 99 & 67.4 (+0.5) & 64 & 0.78 & 0.72 \\
\cellcolor[HTML]{B0B3B2}                           & VGG13     & 70.2 & 99 & 71 (+0.8)   & 68 & 0.76 & 0.77 \\
\cellcolor[HTML]{B0B3B2}                           & VGG16     & 70.2 & 99 & 71.4 (+1.2) & 62 & 0.61 & 0.71 \\
\cellcolor[HTML]{B0B3B2}                           & VGG19     & 70.2 & 99 & 71.8 (+1.6) & 51 & 0.38 & 0.68 \\
\cellcolor[HTML]{B0B3B2}                           & ResNet18  & 66.0 & 96 & 67.8 (+1.8) & 90 & 0.98 & 0.99 \\
\cellcolor[HTML]{B0B3B2}                           & ResNet50  & 71.4 & 96 & 73.5 (+2.1) & 80 & 0.93 & 0.92 \\
\multirow{-7}{*}{\cellcolor[HTML]{B0B3B2}\rotatebox[origin=c]{90}{CIFAR100}} & ResNet101 & 72.2 & 96 & 73.5 (+1.3) & 63 & 0.87 & 0.77 \\ \hline
\end{tabular}
\end{table*}

\subsection{Pretraining on ViTB-16}
Here, we present the results of analyzing VitB-16 architecture trained from scratch on ImageNet and finetuned from a model pretrained in a self-supervised fashion \cite{swag}. All results correspond to networks decomposed and finetuned to respect the rank found from binary search.
\begin{table*}[h]
\small
\caption{Results for Utilized Rank Decomposition for ViTB-16 trained with and without self supervised training \cite{swag} $\dagger$ The increase in accuracy for linear models after finetuning  with decomposed layers is an unfair comparison since the original network only finetuned the linear head.}
\label{tab:imnet_pretraining}
\small
\addtolength{\tabcolsep}{1.5pt}
\def\arraystretch{1.25}%
\begin{tabular}{|ccccccc|}
\hline
\rowcolor[HTML]{B0B3B2} 
 &
  \begin{tabular}[c]{@{}c@{}}Orig Acc\\  (\%)\end{tabular} &
  \begin{tabular}[c]{@{}c@{}}Orig ALU\\  (\%)\end{tabular} &
  \begin{tabular}[c]{@{}c@{}}Proj Acc\\  (\%) ($\Delta$)\end{tabular} &
  \begin{tabular}[c]{@{}c@{}}True ALU\\  (\%)\end{tabular} &
  \begin{tabular}[c]{@{}c@{}}Params\\  Ratio\end{tabular} &
  \begin{tabular}[c]{@{}c@{}}Flops\\  Ratio\end{tabular} \\ \hline
\cellcolor[HTML]{D4D4D4}ViTB16 - scratch & 80.9 & 94 & 80.7 (-0.2) & 35 & 0.48 & 0.33 \\
\cellcolor[HTML]{D4D4D4}ViTB16, FT, Lin  & 81.7 & 97 & 84.0$^\dagger$       & 69 & 0.87 & 0.74 \\
\cellcolor[HTML]{D4D4D4}ViTB16, FT, E2E  & 85.3 & 97 & 85.1 (-0.2) & 57 & 0.79 & 0.54 \\ \hline
\end{tabular}
\end{table*}

\subsection{ViTB-16 with different accuracy drop tolerance, $\mathbf{\epsilon}$}
Here, we present the results of analyzing ViTB-16 architecture trained from scratch with varying accuracy drop tolerance per layer, per transformation.  All results correspond to networks decomposed and finetuned to respect the rank found from binary search.
\begin{table}[H]
\centering
\small
\caption{ViTB-16 pretrained network from torchvision, analyzed for dimensions with varying $\epsilon$ (percentage accuracy drop tolerance per transformation per layer ).}
\label{tab:diffdrop}
\addtolength{\tabcolsep}{2pt}
\def\arraystretch{1.5}%
\begin{tabular}{|
>{\columncolor[HTML]{D4D4D4}}l cccc|}
\hline
\cellcolor[HTML]{B0B3B2} &
  \multicolumn{1}{l}{\cellcolor[HTML]{B0B3B2}\begin{tabular}[c]{@{}l@{}}Acc\\  (\%)\end{tabular}} &
  \multicolumn{1}{l}{\cellcolor[HTML]{B0B3B2}\begin{tabular}[c]{@{}l@{}}ALU\\  (\%)\end{tabular}} &
  \multicolumn{1}{l}{\cellcolor[HTML]{B0B3B2}\begin{tabular}[c]{@{}l@{}}Params\\  Ratio\end{tabular}} &
  \multicolumn{1}{l|}{\cellcolor[HTML]{B0B3B2}\begin{tabular}[c]{@{}l@{}}Flops\\  Ratio\end{tabular}} \\ \hline
ViTB16: Original                      & 80.9 & 93.8 & 1.00 & 1.00 \\
ViTB16: $\epsilon$ = 0.01 & 81.2 & 57.9 & 0.76 & 0.58 \\
ViTB16: $\epsilon$ = 0.05 & 80.8 & 38.7 & 0.54 & 0.37 \\
ViTB16: $\epsilon$ = 0.1  & 80.7 & 34.6 & 0.48 & 0.33 \\
ViTB16: $\epsilon$ = 0.5  & 79.9 & 22.5 & 0.31 & 0.20 \\ \hline
\end{tabular}
\end{table}

\subsection{Literature Review}
\label{sec:litrev}

Determining the rank of learning subspaces has garnered a lot of interest due to its theoretical implications on capacity and generalization of neural networks and its application to model compression. Theoretically rigorous works that find the rank of learning subspaces often show results on small networks and are unable to scale due to computational intractability \cite{intrinsicdimensionlandscape, genguarantee}. On large-scale networks, the low rank nature is assumed and empirically shown to give good results. Many of them exploit the low rank nature of weight matrices to reduce the number of parameters in neural networks by factoring the learned weights in each layer into products of low rank matrices \cite{, rankcollapse, rankdiminish, lowrankbias, decompsoenet, lowrankexpansion, lowranktrain, optimaldecompose, compacter, Garg2020, luca, Zhang2014}.  

Different approaches define \textit{intrinsic} rank differently. Most works find the rank of the weight matrices using matrix factorizations like the SVD \cite{decompsoenet, lowrankexpansion}. Some works constrain this rank statically based on the singular values of the $W$ matrix \cite{rankpca,lowrankinit,lowranktrain,optimaldecompose,lowrankbias,kim2019efficient}, while others learn the rank as part of the optimization procedure \cite{ranklearning, lowranklth, wen2017coordinating}. Rather than predefining the rank via a factorization, another technique that has been used is to construct approximate low rank projection matrices by leveraging the distributional Johnson-Lindenstrauss lemma \cite{jl, linformer} via random projections. These approaches differ from ours in that we project our weights onto the subspaces produced from the input and output activations, which is a architecture-dependent \textit{and} data-dependent approach. Low rank projection based approaches have been applied to transformers \cite{attention} in the past by projecting the weights onto a low rank subspace such as in Linformer \cite{linformer} which sought to reduce the $O(n^2)$ self-attention complexity or SliceGPT \cite{slicegpt} which uses PCA projections to prune large language models.

LoRA \cite{lora} has become the de-facto standard of finetuning large models on downstream tasks. It assumes that the weight updates are low rank and can be restricted to a low-dimensional subspace. It does not restrict the rank of the final, fused weights. We differ from LoRA in that we study and limit the rank of the weights. Our formulation remains compatible with LoRA finetuning on downstream tasks.
In parallel, there are many works that achieve efficiency by quantization, pruning, and knowledge distillation \cite{gptq, sparsegpt, kd, transformer_kd}. In this work, we focus on efficiency via low rank decompositions, and expect that our resulting networks to remain compatible with many of these techniques.

\end{document}